\documentclass[conference,10pt]{IEEEtran}
\IEEEoverridecommandlockouts

\usepackage{cite}
\usepackage{amsmath,amssymb,amsfonts}
\usepackage{graphicx}
\usepackage{textcomp}
\usepackage{xcolor}
\usepackage[linesnumbered,ruled,vlined]{algorithm2e}
\usepackage{algpseudocode}
\usepackage{amsmath}
\usepackage{caption}
\usepackage{graphicx}
\usepackage{float} 
\usepackage{subfigure}
\usepackage{subcaption}
\def\BibTeX{{\rm B\kern-.05em{\sc i\kern-.025em b}\kern-.08em
    T\kern-.1667em\lower.7ex\hbox{E}\kern-.125emX}}

\begin{document}

	\title{Enhanced SPS Velocity-Adaptive Scheme: Access Fairness in 5G NR V2I Networks\\

	}
	
	\author{
		
		Xiao Xu\textsuperscript{1,2}, Qiong Wu\textsuperscript{1,2,*}, Pingyi Fan\textsuperscript{3}, and Kezhi Wang\textsuperscript{4} \\
		\textsuperscript{1}\textit{School of Internet of Things Engineering, Jiangnan University, Wuxi 214122, China} \\
		\textsuperscript{2}\textit{ School of Information  Engineering, Jiangxi Provincial Key Laboratory of Advanced Signal Processing } \\
		\textit{and Intelligent Communications, Nanchang University, Nanchang 330031, China} \\
		\textsuperscript{3}\textit{Department of Electronic Engineering, State Key laboratory of Space Network and Communications,Beijing } \\
		\textit{National Research Center for Information Science and Technology, Tsinghua University, Beijing 100084, China} \\
		\textsuperscript{4}\textit{Department of Computer Science, Brunel University, London, Middlesex UB8 3PH, U.K} \\
		
		Email: xuxiao@stu.jiangnan.edu.cn,\textsuperscript{*}qiongwu@jiangnan.edu.cn, fpy@tsinghua.edu.cn, Kezhi.Wang@brunel.ac.uk \\

	}
	
	\maketitle
	
	\begin{abstract}
		Vehicle-to-Infrastructure (V2I) technology enables information exchange between vehicles and road infrastructure. Specifically, when a vehicle approaches a roadside unit (RSU), it can exchange information with the RSU to obtain accurate data that assists in driving. As the 3rd Generation Partnership Project (3GPP) Release 16, which includes the 5G New Radio (NR) Vehicle-to-Everything (V2X) standards, vehicles typically adopt mode-2 communication using sensing-based semi-persistent scheduling (SPS). By this approach, vehicles identify  resources through a selection window and exclude ineligible resources based on information from a sensing window. However, vehicles often drive at different speeds, resulting in varying amounts of data transmission with RSUs as they pass by, which leads to unfair access. Therefore, developing an scheme that accounts for different vehicle speeds to achieve fair access across the network is essential. This paper formulates an optimization problem for vehicular environment and proposes a multi-objective optimization scheme to address it by adjusting the selection window size. Experimental results validate the efficiency of the proposed method.
	\end{abstract}
	
	
	\section{Introduction}
	
	\label{sec1}
	With the 3rd Generation Partnership Project (3GPP) Release 16, the initial Vehicle-to-Everything (V2X) standard grounded in the 5G New Radio (NR) was introduced as a supplement to Long-Term Evolution (LTE) V2X communication. NR V2X supports two types of communication: mode-1 (centralized) and mode-2 (distributed) \cite{1,26}.
	Vehicles may operate within network coverage in mode-1, and resources are scheduled through the Base station(BS). In contrast, mode-2 allows vehicles to autonomously allocate resources with the sensing-based semi-persistent scheduling (SPS) mechanism, significantly enhancing the flexibility of resource scheduling for vehicles.
	Modern vehicles are often outfitted with cameras, LiDARs, and other sensors to sense the environment\cite{2,3}. However, due to limited onboard computational resources, it is challenging for vehicles to process the large-scale data they produce \cite{23,27}. To address this,  Vehicle-to-Infrastructure (V2I) technology is applied in vehicular network scenarios to obtain real-time data that assists with driving\cite{4,22}.
	In NR V2I mode-2, vehicles typically employ sensing-based SPS mechanisms for resource scheduling\cite{5}. However, vehicles on different lanes often travel at varying speeds, resulting in different durations of time spent within an RSU's coverage area. Furthermore, transmission failures may occur, leading to an unequal amount of successfully transmitted data for vehicles of different speeds\cite{24,25}. This difference, referred to as unfair access, often results in high-speed vehicles receiving less information than low-speed vehicles, which increases the likelihood of incorrect decisions and safety risks.
	
	In summary, designing an access scheme that ensures fair data access in vehicular networks is of great importance. As far as we know, there has been no prior research specifically addressing data access fairness in 5G NR V2I, which motivates our work.
	
	The contributions of this work are summarized :
	
	\begin{itemize}
		\item[1)] We propose a speed-adaptive selection window adjustment scheme for sensing-based SPS scheduling in 5G NR V2I mode 2 to achieve fair network access.
		\item[2)] We define a fairness index to represent the fairness of data access for vehicles at different speeds, proving that this index is a function of both speed and the selection window size.
		
	\end{itemize}
	
	\begin{figure*}[h]
		\centering
		\includegraphics[width=18cm, scale=0.7]{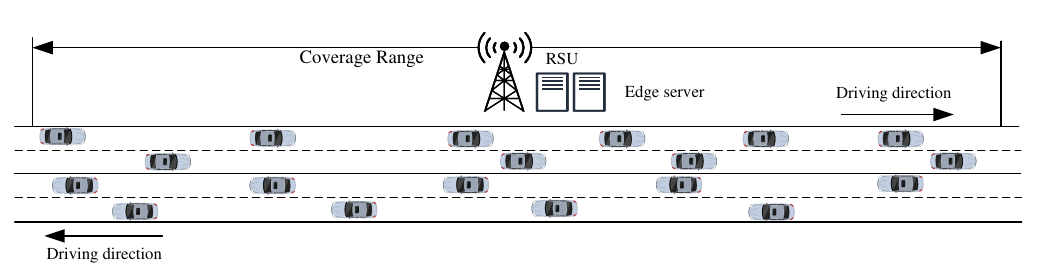}
		\caption{System Model}
		\label{fig1}
		\vspace{-0.2cm}
	\end{figure*}
	The structure of the rest of this work is outlined below:
	Section \ref{sec2} provides an overview of related works. Section \ref{sec3} presents the system model. Section \ref{sec4} presents the fairness index. Section \ref{sec5} formulates the optimization problem and describes solution using NSGA-II. Section \ref{sec6} details the experimental results. Section \ref{sec7} concludes the paper.

	\section{Related Work}
	\label{sec2}
	In this section, we first introduce the relevant improvements related to SPS, followed by a discussion on the enhancements made to network fairness.
	
	\subsection{Sensing Based Semi-Persistent Scheduling}
	Some studies have been worked on the SPS within NR V2X Mode-2. 
	In \cite{6}, Daw \textit{et al.} proposed a priority-based SPS scheme that categorizes emergency vehicles separately and introduced a complementary probabilistic collision mitigation mechanism to minimize the collision probability for high-priority vehicles in the network. 
	In \cite{7}, Jeon \textit{et al.} significantly reduced data packet conflicts in the C-V2X mode-4 by minimizing the uncertainty in resource selection. Specifically, they utilized a “lookahead” approach to eliminate collisions caused by unawareness of other users' decisions. 
	In \cite{20}, Dayal \textit{et al.} extended SPS to accommodate adaptive RRI, referred to as SPS++, to address the severe underutilization and overutilization of radio resources in various vehicular traffic scenarios. This issue significantly impairs the timely dissemination of BSMs, thereby increasing the risk of collisions.
	In \cite{21}, Gu \textit{et al.} proposed an SPS analytical model that quantifies the impact of beacon rate, range settings, and system configuration on the probability of access collisions and delay outage. The analytical model provides critical insights and guidance for adapting and optimizing protocol parameters, including sensing range, transmission power, and resource reservation.
	
	The studies above have improved the SPS mechanism from various perspectives. However, none of these works have considered leveraging the sensing window size to address the issue of access fairness in vehicular networks.
	
	\subsection{Fairness of network}   
	Several studies have proposed solutions to address the fair access of vehicle caused by varying vehicle speeds. In \cite{9}, Wan \textit{et al.} proposed modifying the contention window in the IEEE 802.11p protocol to address vehicle access fairness. In \cite{11}, Wu \textit{et al.} addressed vehicle access fairness in a platooning scenario by dynamically tuning the minimum contention window according to vehicle speeds under the IEEE 802.11p. 
	In \cite{18}, Praghash \textit{et al.} proposed a metric to achieve a certain level of fairness among network users and employed a reinforcement learning (RL) algorithm to mitigate conflicts between clients.
	In \cite{19}, Song \textit{et al.} proposed a novel two-phase scheme, termed Energy-aware UAV Relay Transmission (EURT), where the users' limited residual energy (RE) leads to unfair transmission delays within the network. The scheme aims to balance users' transmission delays and network fairness.

	However, the studies above have failed to simultaneously consider the issue of fair access in vehicular networks, particularly under the 5G NR V2X protocol. Only a few works have addressed fair access under the IEEE 802.11p protocol. Moreover, since the current SPS mechanism cannot flexibly adjust the selection window size based on vehicle speed, it fails to dynamically maintain V2I access fairness according to vehicle speed. Consequently, high-speed vehicles encounter difficulties in exchanging sufficient information with the RSU.
	Therefore, we are motivated to conduct this study.

	\section{System Model}
	\label{sec3}	
	\subsection{Scene Model}
	\begin{figure}[htbp]
		\centering
		\includegraphics[width=8.5cm, scale=0.8]{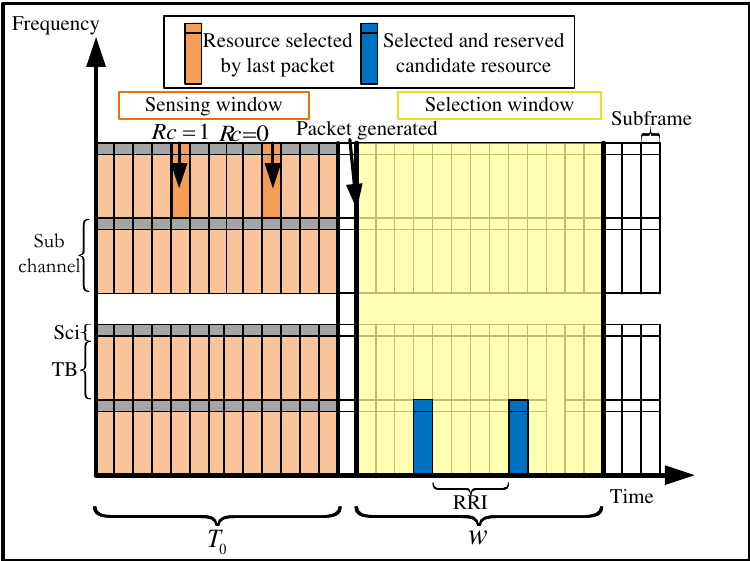}
		\caption{SPS Model}
		\label{fig1}
	\end{figure}
	As depicted in Fig.1, we consider a highway environment featuring $N$ lanes, with an RSU deployed along the roadside as an edge server. The lanes are divided into two directions.
	vehicles arrive within the RSU following a Poisson process. Once a vehicle enters the RSU's communication range, it will undertake information transmission with the RSU and extract useful information.

	\subsection{Sensing Based Semi-Persistent Scheduling}
	Each vehicle adopts 5G NR V2X mode 2 for data transmission and undergoes resource allocation through the SPS. 
	
	~\
	
	Specifically, as Fig.2, the channel is partitioned into subframes, each being 1ms. Within the frequency spectrum, the channel is decomposed into subchannels, each composed of multiple consecutive resource blocks (RBs). A subchannel can be divided into two parts: the side link control information (SCI), occupying two RBs, and the transmission block (TB), occupying the remainder. One subchannel combined with one subframe is referred to as a physical resource block (PRB). Upon reserving a PRB, it generates a reselection counter (RC) randomly and decrements it by 1 after transmission. When $RC = 0$, it will reselect resources with a probability of $ 1 - P$, otherwise, it will continue to utilize the previous resources.
	During the reselection process, vehicle initially identifies resources within the selection window, where the selection window's size $w$ is determined independently by the vehicle. Subsequently, the vehicle determines the resources that should be excluded, namely, according to the information provided by the perception window of size 1000ms.

	\section{Fairness Index}
	\label{sec4}	
	This section focuses on deriving the correlation between the fairness metric for vehicles at different speeds and the parameters of speed and selection window size. Additionally, we define a fairness index to quantify the network's fairness.
	
	\subsection{Transmission rate}		
	Achieving system fairness implies that vehicles traveling at different speeds should transmit an equal average amount of data while within the coverage area of the edge device. Therefore, we can express this as:
	\begin{equation}
		{E[Bit^{i}]=C},
		\label{eq1}
	\end{equation}
	where $E[\cdot]$ is the averaged operation. $Bit^{i}$ represents the amount of data transmitted by vehicle $i$ within the base station's range. $C$ is a constant, and since transmission may fail with a certain probability, we consider the expected value. Specifically, $Bit^{i}$ can be expressed as:
	\begin{equation}
		{Bit^{i}={{C}_{i}}\cdot {{T}_{i}} },
		\label{eq2}
	\end{equation}
	Therefore, Eq.\eqref{eq1} can be rewritten as:
	\begin{equation}
		{{C}_{i}}\cdot {{T}_{i}}\cdot {{P}_{PRR}}^{_{i}}=C,
		\label{eq3}
	\end{equation}
	where ${C}_{i}$ represents the transmission rate of vehicle $i$, and ${T}_{i}$ denotes the time vehicle $i$ spends within the BS's range. ${{P}_{PRR}}^{_{i}}$ is the probability of successful transmission. Therefore, ${T}_{i}$ can be expressed as:
	\begin{equation}
		{{T}_{i}}=\frac{R}{{{v}_{i}}},
		\label{eq4}
	\end{equation}
	where $R$ represents the coverage range of the RSU, and ${v}_{i}$ denotes the speed of vehicle.
	 	Based on Shannon's theorem, ${C}_{i}$ can be expressed as:
	\vspace{-0.1cm}
	\begin{equation}
		{{C}_{i}}=B{{\log }_{2}}(1+\frac{{{p}_{i}}\cdot {{h}_{i}}(t)\cdot {{({{d}_{i}}(t))}^{-\partial }}}{{{\sigma }^{2}}}),
		\label{eq5}
	\end{equation}
	
	~\	
	
	~\	
	
	$B$ represents the bandwidth, ${p}_{i}$ is the transmission power. ${h}_{i}(t)$ represents the channel gain.
	${d}_{i}(t)$ is the distance between vehicle $i$ and the BS, which depends on the vehicle's speed. ${\partial }$ denotes the path loss exponent.  ${\sigma }^{2}$ and is the noise power. The distance ${d}_{i}(t)$ can be described as:
	\begin{equation}
		{{d}_{i}}(t)=\left\| {{P}_{o}}^{i}-{{P}_{o}}^{B} \right\|,
		\label{eq6}
	\end{equation}
	where ${P}_{o}^{i}$ represents the position of vehicle $i$, and ${P}_{o}^{B}$ is the position of the BS. The ${P}_{o}^{i}$ can be described as:
	\begin{equation}
		{{P}_{o}}^{i}(t)=({{v}_{i}}t,0,0).
		\label{eq7}	
	\end{equation}
	
	According to \cite{12}, we adopt an autoregressive (AR) model to express the correlation between ${h}_{i}(t)$ and ${h}_{i}(t-1)$:
	\begin{equation}
		{{h}_{i}}(t)={{\rho }_{i}}{{h}_{i}}(t-1)+e(t)\sqrt{1-\rho _{i}^{2}},
		\label{eq8}	
	\end{equation}
	where ${\rho }_{i}$ represents the channel correlation coefficient during successive time intervals, and $e(t)$ is a complex Gaussian random error vector. Considering the mobility of the vehicle, which introduces Doppler effects. we use Jake’s fading spectrum, ${{\rho }_{i}}={{J}_{0}}(2\pi f_{d}^{i}t)$, where ${{J}_{0}}(\cdot )$ is the zeroth-order Bessel function of the first kind. $f_{d}^{i}t$ represents vehicle $i$'s Doppler frequency. The Doppler shift can be expressed as:

	\begin{equation}
		f_{d}^{i}=\frac{v_i}{{{\Lambda }_{0}}}\cos \theta ,
		\label{eq9}	
	\end{equation}
	where ${\Lambda }_{0}$ is the wave length, and $\cos \theta $ is the angle between the communication direction and the direction of motion.
	
	\subsection{Successful decoding probability}
	Next, we will analyze ${{P}_{PRR}}^{_{i}}$, which represents the probability that the data packet transmitted by vehicle $i$ is successfully decoded by the BS:
	
	\begin{equation}
		{{P}_{PRR}}^{_{i}}=\prod\limits_{j\ne \text{i}}{(1-{{\delta }^{j}}_{COL})\cdot }\prod\limits_{j\ne \text{i}}{(1-{{\delta }^{j}}_{HD})},
		\label{eq10}	
	\end{equation}
	where ${{\delta }^{j}}_{COL} $ represents the data packet collisions' probability between vehicle $i$ and vehicle $j$.  When several vehicles nearly simultaneously attempt to select resources, there is a possibility that they may choose the same PRB. Based on the model in \cite{13}, ${{\delta }^{j}}_{COL} $ can be expressed as:
	\begin{equation}
		{{\delta }^{j}}_{COL}={{P}_{O}}{{P}_{SH|O}}\frac{{{C}_{Ca}}}{N_{Ca}^{2}},
		\label{eq11}	
	\end{equation}
	where ${P}_{O}$ is the probability of overlap between the selection windows of vehicle $i$ and vehicle $j$. ${P}_{SH|O}$ is the probability which vehicle $i$ and vehicle $j$ select resources from their shared selection window. $N_{Ca}$ is the average number of candidate PRBs. ${P}_{O}$ and ${P}_{SH|O}$ can be expressed as:
	
	\begin{equation}
		{{P}_{O}}=\frac{{{w}_{i}}+{{w}_{j}}+1}{1000\cdot {{2}^{\mu }}RRI}.
		\label{eq12}	
	\end{equation}
	\begin{equation}
		{{P}_{SH|O}}={{(\frac{{{N}_{Sc}}{{N}_{Sh}}}{{{N}_{r}}})}^{2}},
		\label{eq13}	
	\end{equation}
		\vspace{-0.1cm}
	where ${N}_{Sc}$ is the number of subchannels. ${N}_{Sh}$ is the shared resources number within the overlapped selection window. 
	
	${N}_{Sh}$ can be expressed as:
	\begin{equation}
		{{N}_{Sh}}=\frac{({{w}_{i}}+1)({{w}_{j}}+1)}{{{w}_{i}}+{{w}_{j}}+1}.
		\label{eq14}	
	\end{equation}
	
	${{\delta }^{j}}_{HD}$ represents the probability that vehicles use the same time slot for transmission. Due to the half-duplex nature of vehicle communication, the receiver cannot decode the data packet from the transmitter if both vehicles transmit simultaneously. Based on the model in \cite{13}, this can be expressed as:
	\begin{equation}
		{{\delta }^{j}}_{HD}=\frac{{{\tau }_{j}}}{1000},
		\label{eq15}	
	\end{equation}
	where ${\tau }_{j}$ represents the packet generation frequency of vehicle $j$. Therefore, it is known that ${{P}_{PRR}}^{_{i}}$ is a function of $w$, where $w$ typically refers to the resource selection window.
	\subsection{Fairness index}
	For brevity, we only consider the fairness index at a certain time. Eq.\eqref{eq3} can be further expressed as:
	
	\begin{equation}
		\begin{aligned} 
			C=&B{{\log }_{2}}(1+\frac{{{p}_{i}}\cdot {{h}_{i}}\cdot {{({{d}_{i}})}^{-\partial }}}{{{\sigma }^{2}}})\cdot \frac{R}{{{v}_{i}}}\\ 
			&\cdot \prod\limits_{j\ne \text{i}}{(1-{{\delta }^{j}}_{COL})\cdot }\prod\limits_{j\ne \text{i}}{(1-{{\delta }^{j}}_{HD})}.
			\label{eq16}
		\end{aligned}	
	\end{equation}
	
	We eliminate the items that have no relation to vehicle $i$. Therefore, Eq. \eqref{eq16} can be further expressed as: 
		\begin{equation}
			\begin{aligned} 
				K_{index}^{i}=\frac{C}{{{C}'}}=&{{\log }_{2}}(1+\frac{{{p}_{i}}\cdot {{h}_{i}}\cdot {{({{d}_{i}})}^{-\partial }}}{{{\sigma }^{2}}})\\
				&\cdot \frac{\prod\limits_{j\ne \text{i}}{(1-{{\delta }^{j}}_{COL})}}{{{v}_{i}}},
				\label{eq17}
			\end{aligned}	
		\end{equation}
	Thus, we have derived the fairness index for vehicle $i$. Furthermore, since ${d}_{i}$ is a function of $v_i $ and ${{P}_{PRR}}^{_{i}}$ is a function of $w$, $K_{index}^{i}$ is a function of both $v$ and $w$. 

	Therefore, by knowing the vehicle's speed, we can adaptively adjust the vehicle's selection window based on speed to achieve overall network fairness.
	
	By averaging the speed and window size of all vehicles in the network, we can obtain:
	\begin{equation}
		K_{index}={{\log }_{2}}(1+\frac{{{p}_{i}}{{h}_{i}}{{{{d}_{i}}(\bar{v})}^{-\partial }}}{{{\sigma }^{2}}})\cdot \frac{\prod\limits_{j\ne \text{i}}{(1-{{\delta }^{j}}_{COL}(\bar{w}))}}{\bar{v}}
		\label{eq18}	
	\end{equation}
	where $\bar{v}$ represents the average speed, and $\bar{w}$ represents the network's average window size. The fairness index 
	$K_{index}$ can measure the overall network's fairness. The vehicle is achieving fair access when $K_{index}^{i}$ approaches $K_{index}$.

	\section{Optimization Problem and Solution}
	\label{sec5}
 
	In this section, we formulate a multi-objective optimization problem and employ the NSGA-II algorithm \cite{14}.
	\vspace{-0.01in}
	\subsection{Optimization Objective}
	\addtolength{\topmargin}{0.02in}
	The optimization goal is to adjust vehicle's selection window sizes to ensure that  data transmitted between each vehicle and the RSU is similar. This implies that $K_{index}^i$ approaches the network's fairness index $K_{index}$. Accordingly, the optimization objective functions can be formulated as: 

	\textbf{Objective 1 to $\boldsymbol{N}$:} Minimize the variation between the fairness index on different lanes and network's fairness index.
		\vspace{-0.1cm}
	\begin{equation}
		\begin{aligned}
			\min_{\boldsymbol{w}} \; &\boldsymbol{F}(\boldsymbol{w}) = [ F_{K_{1}}(\boldsymbol{w}), F_{K_{2}}(\boldsymbol {w}),\dots,F_{K_{N}}(\boldsymbol{w})]^T
			\\
			\qquad \qquad & s.t \\
			& w^{LB} \leq w^i  \leq w^{UB}, i \in [1, \ldots, N],
			\label{eq20}
		\end{aligned}
	\end{equation}
	where
	\begin{equation}
		F_{K_i}(\boldsymbol{w}) =  \left|  K_{index}(\boldsymbol{w}) - K_{index}^i(\boldsymbol{w}) \right|,
		\label{eq21}
	\end{equation}
	$\boldsymbol{w}=\{w^1,w^2,...,w^N\}$. $ w^{LB}$ and $w^{UB}$ indicate the minimum and maximum bounds of the selection window sizes.
	 Our aim is to filter the resulting set of Pareto solutions to find the optimal solution.
	
	\subsection{Optimization Solution}

	We use the NSGA-II algorithm \cite{14} to address the optimization problem. Each population consists of $N$ individuals, representing the vehicle selection windows. The population size is $M$. The algorithm is presented in Algorithm 1.

	1) Initialization Phase: We randomly initialize a set of selection windows within the range $[w^{LB}, w^{UB}]$.
	
	2) Iteration Phase:  Firstly, crossover and mutation are performed.
	 Crossover involves exchanging parts of two parent individuals' genes to generate new offspring. Mutation randomly modifies an individual's genes to prevent the algorithm from converging to a local optimum. At this point, the parent and offspring populations are merged into a new population $Q_n$ for the subsequent selection operation. 

	Next, we use $\boldsymbol{v}=\{v_1, v_2, \dots, v_N\}$ to calculate the objective function values $F_K(\boldsymbol{w})$ according to Eq.\eqref{eq21}.
		
	Then we perform non-dominated sorting.  Non-dominated sorting divides the population into multiple levels according to each individual's dominance rank. Thus we derive $\boldsymbol{F}=[F_1, F_2, \dots]$, $F_1$ means the individuals which rank first, and then, crowding distance is calculated. 
	Parent individuals are selected based on non-dominated sorting and crowding distance. 
	The selection prefers individuals that are ranked higher within the same rank. Individuals with larger crowding distances are preferred.

%
%
%
		
	Thus, the merged population $Q_n$ undergoes non-dominated sorting and selection. 
	Finally, $M$ individuals with higher domination rank and larger crowding distances are selected to form the new population $p_{n+1}$.
	
	3) Optimization Phase: The optimal population is selected from the Pareto front. The optimal population must ensure that the difference between the fairness index of each vehicle is smaller than a threshold. 
	
	~\
	
	Meet the above conditions while also minimizing the total objective function values. Thus, the optimal selection window $\boldsymbol{w}^*$ is obtained.
\begin{algorithm}[h]
	\setlength{\textwidth}{0.8\linewidth}  
	\caption{Non-dominated Sorting Genetic Algorithm II}
	\KwIn{$\boldsymbol{v}=\{v_1, v_2, \dots, v_N\}$, $N_{max}$: Maximum number of generations}
	\KwOut{optimal solution $\boldsymbol{w}^*$}	
	Initialize the population $P_0 = \{\boldsymbol{w_1}, \boldsymbol{w_2}, \dots,\boldsymbol{w_M}\}$ within $[w^{LB},w^{UB}]$ \;
	\For{$n = 0$ \textbf{to} $N_{max}$}{
		\textbf{Crossover:} Generate $P_n^C$ from $P_n$ with crossover\;
		\textbf{Mutation:} Apply mutation to $P_n^C$ to produce $P_n^M$\;
		\textbf{Population Merge:} Form $Q_n = P_n \cup P_n^M$\;
		\textbf{Objective function value:} Calculate $F_K(\boldsymbol{w_i})=\{f_1(\boldsymbol{w_i}), f_2(\boldsymbol{w_i}), \dots, f_M(\boldsymbol{w_i})\}$ based on Eq.\eqref{eq21}, for all $\boldsymbol{w_i}\in Q_n$\;
		\textbf{Non-dominated Sorting:} Compute $\boldsymbol{F}=\{F_1, F_2, \dots\}$ for $Q_n$\;		
		\textbf{Crowding Distance:} Compute crowding distance $d(\boldsymbol{w_i})$ for all $\boldsymbol{w_i}\in Q_n$ with $F_K(\boldsymbol{w_i})$\;
		\textbf{Selection:} Select the top $M$ individuals in $Q_n$ as $P_{n+1}$ using $\boldsymbol{F}$ and $d(\boldsymbol{w_i})$\;
	}
	$P_{\text{filtered}} = \boldsymbol{w} \in  P_{n+1}$ : $F_{K_i}(\boldsymbol{w}) \leq \text{Threshold}, \forall i$\;
	$\boldsymbol{w}^* = \arg \min_{\boldsymbol{w} \in P_{\text{filtered}}} \sum_{i=1}^N F_{K_i}(\boldsymbol{w})$\;
\end{algorithm}

	\section{Numerical Simulation And Analysis}
	
	\label{sec6}
	
	In this section, we validate the effectiveness of our proposed scheme.The experimental setup and implementation details can be accessed from the source code repository at https://github.com/qiongwu86/Enhanced-SPS-Velocity-adaptive-Scheme-Access-Fairness-in-5G-NR-V2I-Networks. 
	The scenario is set as a two-way highway with four lanes, where the speed limit is between 20 and 30 m/s, and the speed difference between adjacent lanes is within 4 m/s. The window size limits and time slot length are configured according to the 5G NR V2I Mode 2 settings.

	
	
Fig. 3 presents the performance metrics of the NSGA-II using hypervolume (HV), Inverted Generational Distance (IGD), Generational Distance (GD), and Spacing. The HV measures the extent to which the solution set covers the objective space, progressively increasing and stabilizing, indicating high diversity. GD evaluates the proximity of the obtained solutions to the true Pareto front, with values steadily decreasing and stabilizing, demonstrating effective convergence. IGD considers both diversity and convergence, yielding higher values than GD. Lastly, Spacing measures the uniformity of solution distribution, remaining consistently low and indicating an even spread of solutions.
\begin{figure}[htbp]
	\centering
	\includegraphics[width=\linewidth]{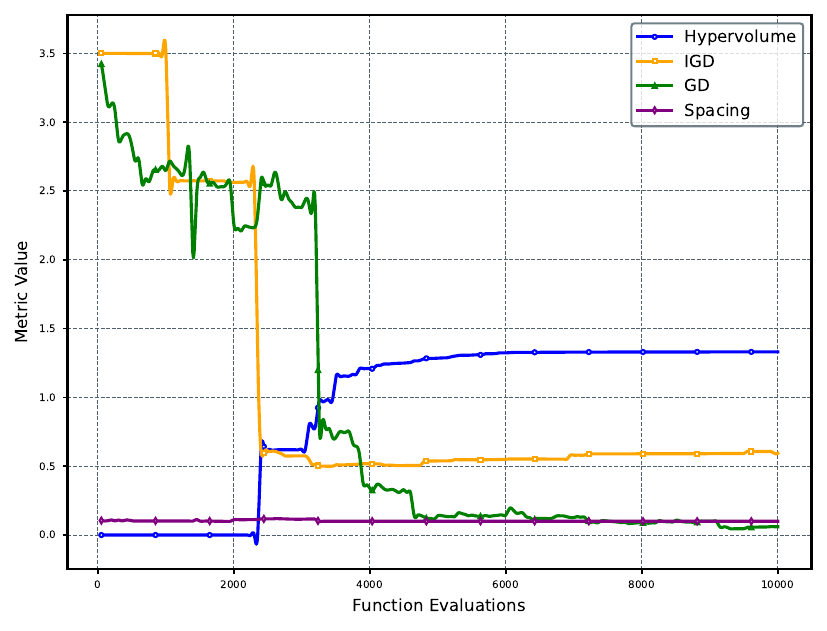}
	\caption{Performance Indicators of the NSGA-II Algorithm}
	\label{fig3}
\end{figure}

\begin{figure}[htbp]
	\centering
	\includegraphics[width=\linewidth]{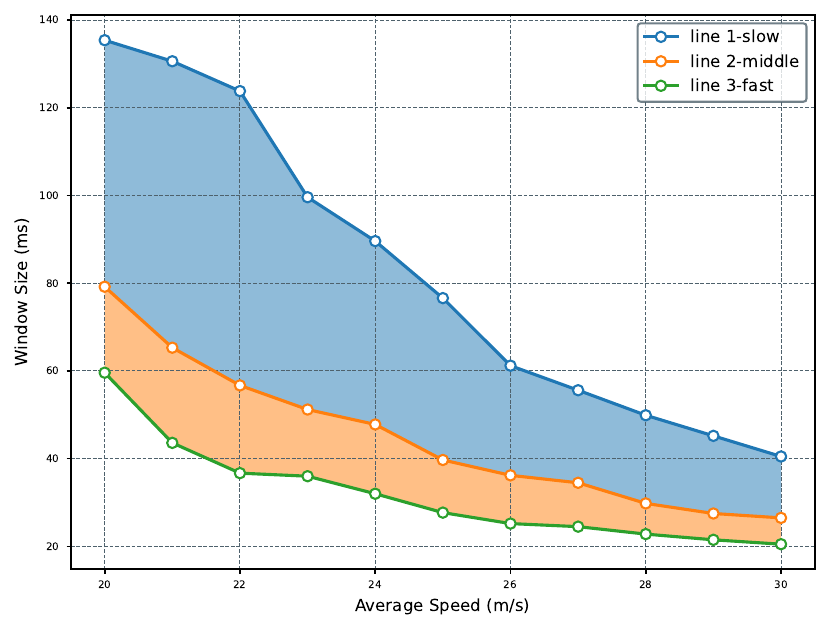}
	\caption{Optimal Selection Window Versus Average Velocity}
	\label{fig4}
\end{figure}

\begin{figure}[htbp]
	\centering
	\includegraphics[width=\linewidth]{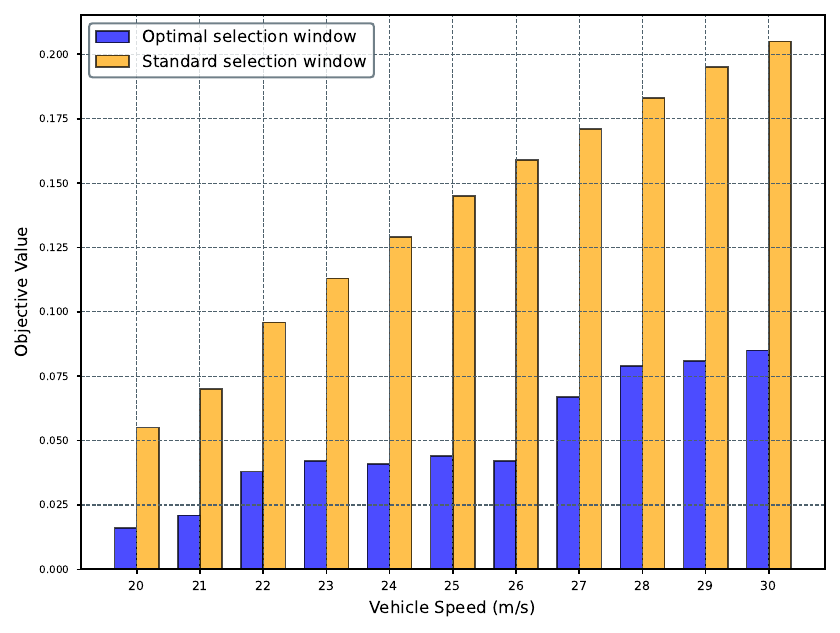}
	\caption{Objective Function Value Versus Average Velocity}
	\label{fig5}
\end{figure}
Fig. 4 illustrates the optimal window size variation for three vehicles traveling in different lanes as the average vehicle speed increases. It can be observed that vehicles with higher speeds tend to have smaller optimal window sizes.
This is because higher speeds reduce the  transmission time with RSU, necessitating a decrease in the window size to maximize data transmission.
Moreover, as the average network speed increases, the window sizes for all vehicles decrease. This is also attributed to the challenge of achieving fairness at higher speeds, which requires a reduction in the window size to balance data transmission across vehicles effectively.

In Fig. 5, we present a comparison of the objective values when vehicles adopt the optimal selection window versus the standard selection window. It can be observed that when using the standard selection window, the objective values of vehicles progressively increase, indicating a gradual loss of fairness in access. This phenomenon arises because faster vehicles will have less staying time in RSU coverage, resulting in transmitting less data, leading to unfairness. However, when the optimal selection window is employed, both the rate of increase and the initial values are significantly smaller compared to the standard window. This improvement is attributed to the vehicles’ ability to adaptively adjust the optimal window based on their speed, thereby striving to maintain fairness.

	\section{Conclusion}
	
	\label{sec7}
	In this paper, we considered the fairness in data access within vehicular networks and proposed a multi-objective optimization method based on adjusting the selection window of the SPS in NR V2I. The goal is to ensure fair access to the base station for vehicles at varying speeds. The simulation results lead to the following conclusions:

	\begin{itemize}
		\item Vehicle speed significantly impacts access fairness. Under the same conditions, higher vehicle speeds result in reduced data exchange with the RSU, making it more challenging to achieve fairness.
		\item The optimal selection window decreases progressively with increasing vehicle speed to reduce data transmission time costs, thereby improving fairness.
	\end{itemize}
	
	In future research, optimization can be broadened to simultaneously consider the minimization of age of information.

	\section*{Acknowledgment}
	This work was supported in part by Jiangxi Province Science and Technology Development Programme under Grant No. 20242BCC32016, in part by the National Natural Science Foundation of China under Grant No. 61701197, in part by the National Key Researchand Development Program of China under Grant No. 2021YFA1000500(4) and in part by the 111 Project under Grant No. B23008. (Corresponding author: Qiong Wu.)

	\ifCLASSOPTIONcaptionsoff
	\newpage
	\fi
	
	\bibliographystyle{IEEEtran}
	\bibliography{IEEEabrv,ref1}

	
\end{document}